\documentclass{article}
\PassOptionsToPackage{numbers, compress}{natbib}

\usepackage[final]{neurips_2024}

\usepackage[utf8]{inputenc} 
\usepackage[T1]{fontenc}    
\usepackage{hyperref}       
\usepackage{url}            
\usepackage{booktabs}       
\usepackage{amsfonts}       
\usepackage{nicefrac}       
\usepackage{microtype}      
\usepackage{xcolor}         
\usepackage{graphicx}
\usepackage{xcolor, soul}
\sethlcolor{pink}
\usepackage{subcaption}
\usepackage{wrapfig}

\title{Morphological Typology in BPE Subword Productivity and Language Modeling}

\author{
    Iñigo Parra\\
    Department of Linguistics\\
    University of California, Berkeley\\
    \texttt{iparra@berkeley.edu} \\
}

\begin{document}

\maketitle

\begin{abstract}
This study investigates the impact of morphological typology on tokenization and language modeling performance. We focus on languages with synthetic and analytical morphological structures and examine their productivity when tokenized using the byte-pair encoding (BPE) algorithm. We compare the performance of models trained with similar amounts of data in different languages. Our experiments reveal that languages with synthetic features exhibit greater subword regularity and productivity with BPE tokenization and achieve better results in language modeling tasks. We also observe that the typological continuum from linguistic theory is reflected in several experiments. These findings suggest a correlation between morphological typology and BPE tokenization efficiency.
\end{abstract}

\section{Introduction}

Since the introduction of the transformer architecture \cite{vaswani2017attention}, large language models (LLMs) have shown unparalleled multilingual performance. Modern generative pretrained transformer (GPT) models are trained on extensive text corpora, typically tokenized using the byte-pair encoding (BPE) algorithm \cite{gage1994bpe,sennrich-etal-2016-neural,radford2019language}. Tokenization is a critical phase in the training process \cite{toraman2023impact}, determining the units the model will predict in an auto-regressive manner.

Morphology is the area of linguistics concerning the study of word formation and structure. It examines how morphemes, the smallest units of meaning in a language, combine to form words. Modern morphological typology distinguishes analytic and synthetic languages. Analytic languages, such as isolating languages, typically have a one-to-one correspondence between words and morphemes, with minimal affixation. Synthetic languages (fusional, agglutinative, and polysynthetic) use inflection and affixation extensively \cite{sapir2004language,greenberg1960quantitative}. These categories form a typological continuum, meaning that most languages exhibit features from multiple types \cite{arkadiev2020morphology}.

It is crucial to investigate whether BPE tokenization is more effective for specific languages due to its widespread use in state-of-the-art LMs. Different morphological structures, such as those in synthetic and analytic languages, pose distinct challenges for tokenization algorithms. Understanding these differences can help optimize tokenization strategies and provide researchers with unique insights into the models' learning process. 

In this study, we analyze the impact of morphological typology on BPE tokenization and language modeling. To address this issue, we ask two questions: (1) \textit{do some morphological typologies condition BPE tokenization?} and, if so, (2) \textit{do language modeling tasks reflect these advantages or disadvantages?} This study compares the effect of BPE on languages with analytic and synthetic morphological features. We perform language modeling, regularity, and productivity experiments and show that:

\begin{itemize}
    \item Synthetic morphology is associated with a \textbf{more productive subword system}
    \item Languages with higher degrees of synthesis show \textbf{lower perplexity and loss}
    \item Synthetic languages achieve \textbf{better generalization faster} when trained in parallel corpora
    \item There is a \textbf{complexity continuum} that favors synthetic languages, reflecting the typological continuum of linguistic theory
\end{itemize}

\begin{figure}[htbp]
    \centering
    \includegraphics[width=0.4\linewidth]{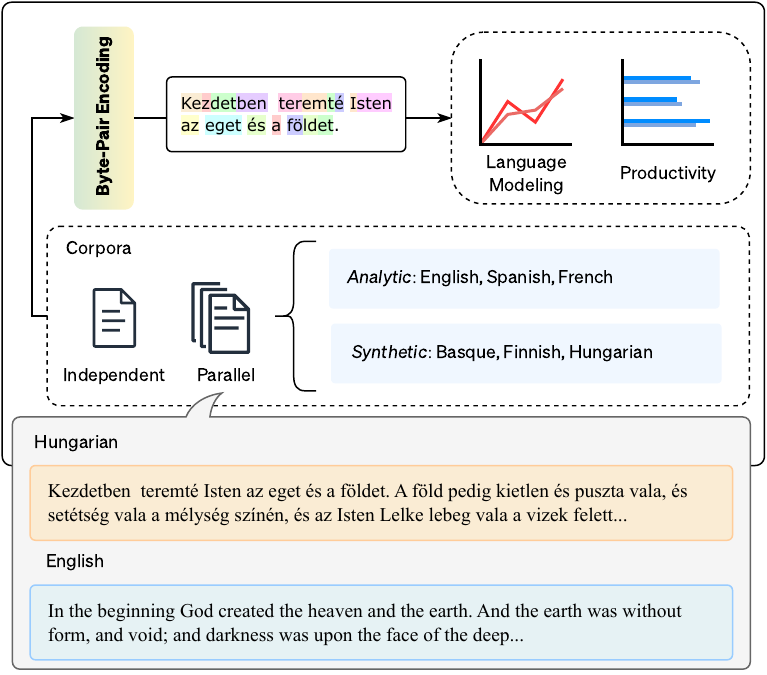}
    \caption{Example of the pipeline. The input consists of parallel and independent corpora. After BPE tokenization, we compare performance on language modeling and compute the subword productivity.}
\end{figure}

\section{Background and Previous Work}

\subsection{Vocabulary}
The choice of the tokenization algorithm is dependent on the task being performed \cite{manohar2023improving}. For language modeling, the algorithm –or tokenizer– converts text data into words and subwords that form its vocabulary. Building the vocabulary involves training a tokenizer to establish the core semantic knowledge of a model. This process is crucial because tokenization can significantly impact the outcome of LMs \cite{rust-etal-2021-good}.

\citet{sennrich-etal-2016-neural} have highlighted the importance of subword tokens, which are crucial for handling rare and unseen words in the training data. For this, subword tokenization techniques such as BPE have been found to be particularly effective. BPE produces merges out of recurrent patterns, which allows it to achieve better generalization compared to other methods like SentencePiece \cite{kudo-2018-subword} or WordPiece \cite{schuster2012japanese}.

As \citet{gutierrez2023languages} mention, BPE tokenization has been classified as irrelevant from a linguistic point of view \cite{galle-2019-investigating,bostrom-durrett-2020-byte,clark2022canine,oncevay-etal-2022-quantifying}. Studies have shown that an increase in linguistically-informed properties does not lead to improved performance in key downstream tasks \cite{domingo2019much,saleva-lignos-2021-effectiveness}. Other authors \cite{mehta2023semantic} have discussed the effectiveness of BPE in minimizing out-of-vocabulary tokens (\texttt{[OOV]} or \texttt{[UNK]}). Reducing \texttt{OOV} tokens can be a significant advantage when dealing with languages with complex morphological typologies. BPE's approach to tokenization ensures that the model retains more information about rare and composite words, which may also cause an increment in the performance and robustness of LMs.

BPE has been previously formalized by \citet{zouhar-etal-2023-formal}. BPE merges sequences $\mu$ that are prominent across the corpus. What BPE tokenization aims to solve is an information compression problem. The compression power of the encoding is given by the reduction of a string $\mu$. For a given string, the original condition is determined by the initial power $G_x(\emptyset) = 0$. Then, the compression preserves monotonicity; as the merges are applied, the power increases\footnote{$\oplus$ is used for concatenation.} (\ref{power_increase}, \ref{recurrence}).

\begin{equation}\label{power_increase}
    G_x(\mu \oplus \mu') \geq G_x(\mu)
\end{equation}
\begin{equation}\label{recurrence}
    G_x(\mu \oplus \mu') - G_x(\mu) \leq G_x(\mu' \oplus \mu'') - G_x(\mu')
\end{equation}

Because BPE is a greedy algorithm aiming to maximize $G$, given \ref{recurrence}, we may say that compression power increases at each step. This increase in compression may continue up to an optimal point, after which further merges are not possible. However, in practice, the halting limit is determined by a vocabulary size $V$.

\subsection{Regularity and Productivity}
Regularity in morphology is defined as the property of forming morphemes according to a set of combinatory rules consistent throughout the language \cite{greenberg1960quantitative}. Since regularity arises from recurrent patterns in different words \cite{bonami2016joint}, regularity does not imply less complex morphological typology. Because of this, we hypothesize that synthetic languages tend to show more regularity because they show more automaticity (i.e., predictability of the forms) \cite{wells1949automatic} due to subword recurrency.

Regularity has also been closely related to productivity \cite{gutierrez2023languages}. This is because both concepts find common ground in frequency: morphemes recurrent throughout many lexemes are also indicators of a productive system \cite{bybee2003phonology,bybee2010language}. Due to this, we expect synthetic languages to display higher frequency, regularity, and productivity.

\subsection{Are All Languages Equally Learnt by LMs?}

According to \citet{chomsky2023noam}, LMs ``are
incapable of distinguishing the possible from the impossible'' (p. 3). This seems to suggest a clear difference between linguistic knowledge and performance. However, it also poses some interesting questions: do some linguistic properties (e.g., morphology) affect language modeling performance? If not, are all languages equally learned by language models (LMs)? 

\citet{cotterell2018all} address this question comparing LSTM and \textit{n}-gram language models trained on translated corpora. They found correlations between morphological richness and performance decrease. However, even if recurrent language models are able to capture complex dependencies \cite{hwang2017character,kawakami2017learning}, attention-based models may reduce these effects. In this line, \citet{koplenig2023languages} analyze LSTM and Transformer models' performance on various languages and found that languages with more speakers were harder to model.

\section{Methodology}

We use six languages to experiment with. To account for the general differences, we divided them into two equal groups representing analytic and synthetic typologies. The analytic group comprised English (primarily analytic), Spanish (primarily fusional), and French (primarily fusional). The synthetic group included Basque (primarily agglutinative with polysynthetic features), Finnish (agglutinative), and Hungarian (primarily agglutinative with polysynthetic features). We also studied the individual differences and variability across languages to check for a possible complexity continuum.

\subsection{Quantifying Productivity}
Given the close relationship between frequency, productivity, and regularity, we began by providing descriptive measurements based on subword usage. We computed the frequency trends of the most used subwords per language to identify initial patterns. After, we tested the significance of these results through continuous sampling and testing.

To address productivity, we expand on previous experiments by adopting the metric proposed by \citet{gutierrez2023languages}. We incorporate minor changes, such as extracting the productivity means per language. After running the BPE algorithm for 300, 400, and 500 merge operations, we computed the productivity means and deviations. This ensured reliable and robust results.

\begin{equation}\label{productivity_eq}
    \rho = \frac{1}{N} \sum_{s \in S} |W_s|
\end{equation}

In Equation \ref{productivity_eq}, $\rho$ represents the productivity of a language. $S$ is the set of all subwords $s$, $W_s$ denotes the set of unique words in which a subword $s$ appears, and $|W_s|$ represents its cardinality (i.e., size). To find the productivity of a language, we compute the average $|W_s|$ determined by its subwords. In other words, we define a productive morphological system as one where subwords appear in many different words. We compute this metric for each language using the Parallel Bible Corpus (PBC) \cite{christodouloupoulos2015massively}.

\subsection{Language Modeling Experiments}
We train six transformer models from scratch for each language modeling experiment. We designed small LMs to balance computational efficiency with experimental rigor. We used four layers and four attention heads, providing sufficient complexity while remaining computationally manageable. We believe this configuration is a reasonable proxy to judge the effect of morphological typology and BPE tokenization on model performance, enabling us to analyze its impact within a constrained resource framework. 

In the first experimental setup, we train models on similar amounts of tokens (100M per language) extracted from the Leipzig Corpora Collection (LCC) \cite{goldhahn-etal-2012-building}. We compare the loss and perplexity throughout the training of each language to provide a comprehensive picture of language learning. We repeat the training process three times per language. In the second, we observe the generalization each language achieves in parallel texts. We train the models on the PBC corpora and compare validation perplexity and loss across languages. We use the same parameters and model architecture used in the previous experiment.

After training, we calculate the mean values of the final metrics for each language within both groups. We then use these means to compute the categories' performance difference ($\Delta$). Additionally, we calculate and compare the standard deviation ($\sigma$) for both groups to assess the robustness of performance across languages within each typological class.

\section{Results}

\subsection{Regularity and Productivity}

\begin{wrapfigure}{r}{0.4\linewidth}
    \centering
    \includegraphics[width=\linewidth]{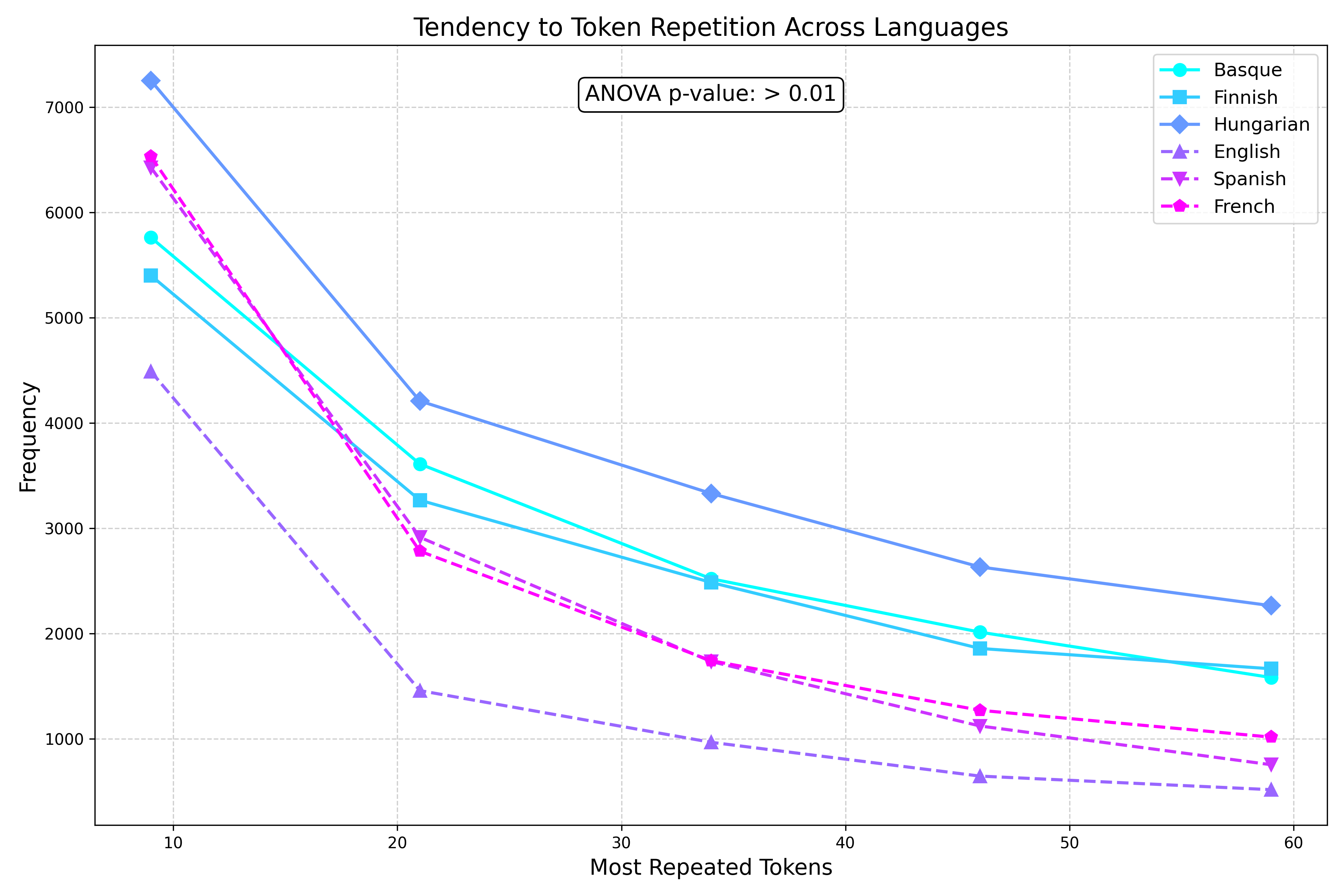}
    \caption{Trends of subword repetition. As the sample increases, the lines form two groups that show distinct behaviors. Red lines represent synthetic languages; green lines represent analytic languages.}
    \label{trends}
\end{wrapfigure}

Our analysis of subword patterns revealed interesting differences between languages. Initially, no distinctive patterns were observed. However, as additional subwords were analyzed and the sample size increased some languages exhibited stabilization, while others showed a sharper decline (Figures \ref{trends} and \ref{top_tokens}). These differences reflect both, (1) that the groupings –synthetic and analytic– seem to behave similarly, and (2) that there is a continuum favoring the richer morphological typologies.

Unpaired $t$-tests comparing the analytic and synthetic groups corroborated the observed visual trends. The tests were corrected using Bonferroni correction. Statistical results supported the hypothesis that synthetic languages exhibit higher automaticity ($p=0.01$). These patterns in subword usage align with theoretical expectations of linguistic structure, which BPE may exploit. Apart from comparing the typological groupings, we also compare the differences between languages using a one-way ANOVA \ref{trends}. The differences observed in the graph were statistically significant ($p=0.01$), which corroborated the visual continuum.

\begin{figure}[htbp]
    \centering
    \includegraphics[width=0.4\linewidth]{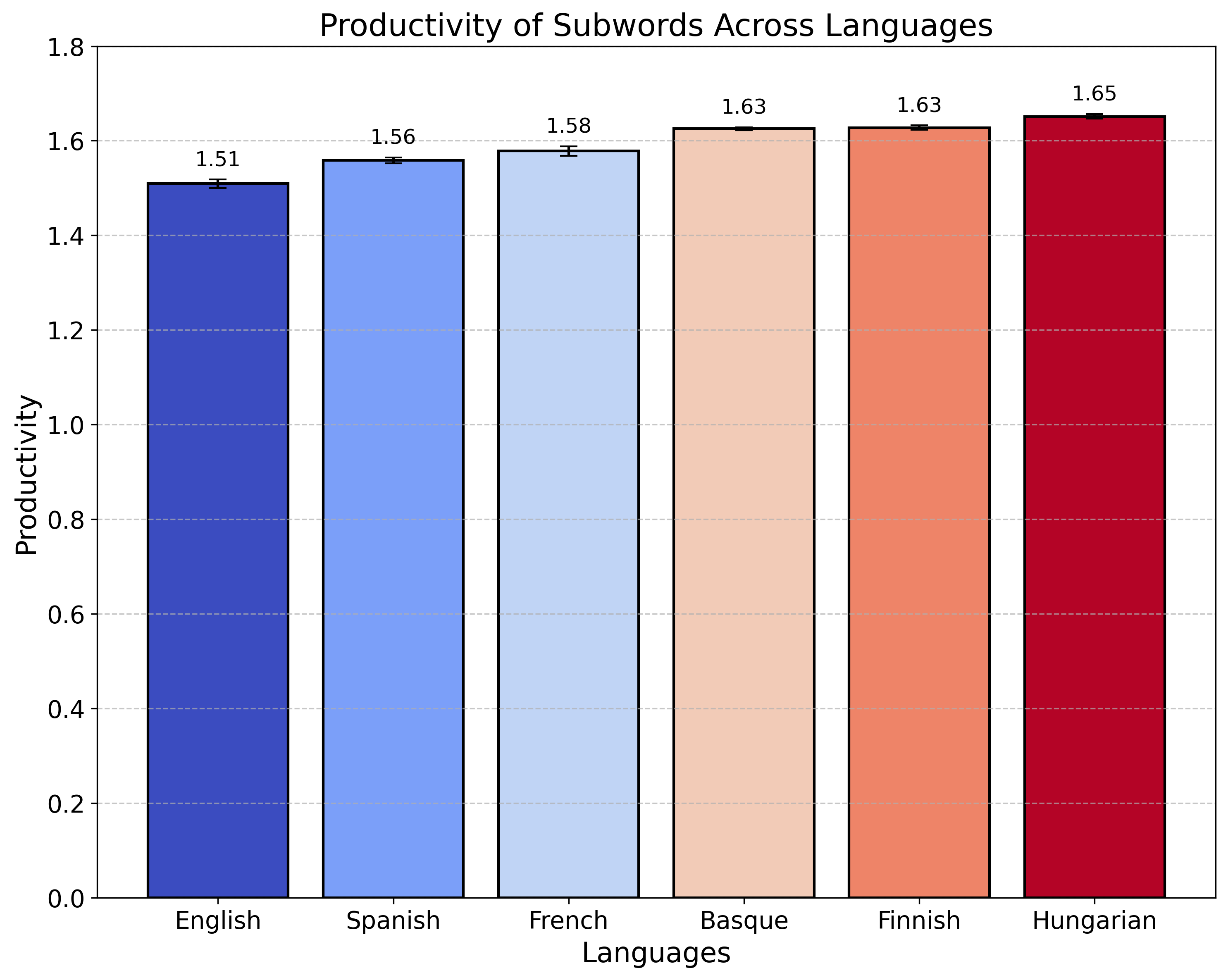}
    \caption{Productivity scores per language after averaging results for 300, 400, and 500 merge operations. Measurements were performed using the PBC parallel corpora. The error bars indicate the standard deviation between rounds of merge operations.}
    \label{productivity}
\end{figure}

The implications of these results extended to productivity scores (see Figure \ref{productivity}). All analytic languages displayed lower productivity compared to the synthetic ones. The reliance of analytic languages on word order and function words results in fewer unique subwords, thereby reducing their scores. This coincides with the results shown in Figure \ref{trends}: the languages that show the highest repetition tendencies are also the most productive.

Basque and Finnish, which are characterized by their agglutinative nature, showed similar productivity scores. Hungarian exhibited a moderately lower score. These differences underscore the nuances within synthetic languages, highlighting that while they generally exhibit higher productivity, there are variations based on specific linguistic characteristics.

Because productivity was determined by the number of unique words in which each subword appeared, high productivity scores were also indicators of regularity. This was expected since synthetic languages tend to show more automaticity. Overall, BPE tokenizers are able to exploit this and make synthetic languages more efficient. We expected this to be a relevant factor in language modeling.

\subsection{Token Frequency}

Another interesting component of the frequency is related to how this evolves as more tokens are analyzed. We computed the rates of change given by the slope for the frequency decay in all language types using ordinary least square regressions. The results in Table \ref{slopes} are based on an analysis of the 100 most frequently occurring subwords. We show clear distinctions between synthetic and analytic languages and between languages themselves. 

\begin{table}[h]
    \centering
    \begin{tabular}{lllll}
    \toprule
        \textbf{Language}\hspace{1cm} & \textbf{Type/Features}\hspace{1cm}            & \textbf{Change/Slope} & $r$     & $R^2$     \\
        \midrule             
        Basque                        & primarily agglutinative + polysynthetic       & -0.70 $\downarrow$    & -0.99    & 0.98     \\
        Finnish                       & primarily agglutinative                       & -0.66 $\downarrow$    & -0.99    & 0.99     \\
        Hungarian                     & primarily agglutinative, + polysynthetic      & -0.66 $\downarrow$    & -0.99    & 0.98     \\
        \midrule
        English                       & primarily analytic                            & -1.03 $\uparrow$      & -0.99   & 0.98     \\
        Spanish                       & primarily fusional, + analytic                & -1.02 $\uparrow$      & -0.98   & 0.97     \\
        French                        & primarily fusional, + analytic                & -0.95 $\uparrow$      & -0.98   & 0.97     \\
    \bottomrule
    \end{tabular}
    \caption{Results of the rate of change (provided by the slope) on most repeated subwords for analytic and synthetic languages using the top 100 subwords. $\downarrow$ is best. Since the repetition decreases, we expected the $r$ values to be negative.}
    \label{slopes}
\end{table}

\begin{figure*}[htbp]
    \centering
    \includegraphics[width=1\linewidth]{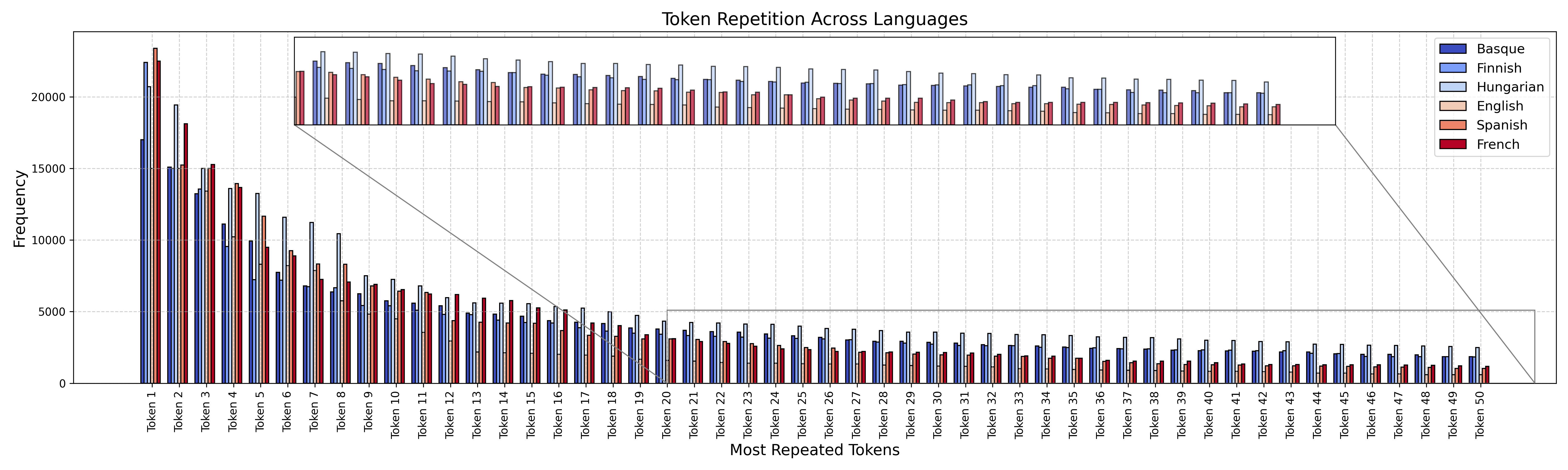}
    \caption{Frequencies of the top $n$-th most repeated subword. As observed in the graph, as further tokens are analyzed, the tokens in synthetic languages show higher frequencies.}
    \label{top_tokens}
\end{figure*}

For any number of top subwords, the rate of change of the frequency function with respect to the subword rank is greater for analytic languages than for synthetic languages. In other words, as we move from the most frequent subwords to the less frequent ones, the decay in frequency is faster in analytic languages than in synthetic languages. We formalize this in Equation \ref{formalism}. Let $f^A(i)$ and $f^S(i)$ be the frequencies of the $i$-th most repeated subword in analytic and synthetic languages, respectively. Let $k$ be any positive integer representing the subword rank under consideration. Then:

\begin{equation}\label{formalism}
\forall k \in \mathbb{Z}^+, \quad f^A(k) - f^A(k+1) > f^S(k) - f^S(k+1)
\end{equation}

Equation \ref{formalism} is supported by visual evidence shown in Figures \ref{trends}, \ref{productivity}, and \ref{top_tokens}, and the statistical results from the $t$-tests, ANOVA, and the slopes in Table \ref{slopes}

\subsection{Language Modeling}

We used the general Transformer architecture with fewer hyperparameters to make the experimentation more manageable and efficient. Instead of using six layers and six heads, we reduced both to four. We used a 0.2 dropout value to mitigate overfiting. We changed the Adam optimizer for AdamW with a 1e-4 learning rate, $\beta_1$ set to 0.9, and $\beta_2$ to 0.98.

Given our study's computational constraints and exploratory nature, we opted for a more compact model configuration, particularly in terms of embedding dimensions. While larger models with embedding dimensions of 512, 768, or even 1024 have become common in language modeling tasks, we intentionally chose a 256-dimensional embedding space for several reasons. Our primary research question was the impact of morphological typology on tokenization and language modeling, not the absolute state-of-the-art performance. A smaller embedding size provided a lower-resolution representation that still captured essential linguistic features while being more sensitive to morphological variations. This controlled setting helped isolate the effects of morphology from the sheer representational power of larger models, which might have masked or overshadowed these nuanced linguistic phenomena.

\begin{figure*}[htbp]
    \centering
    \begin{subfigure}[b]{0.49\textwidth}
        \includegraphics[width=1\linewidth]{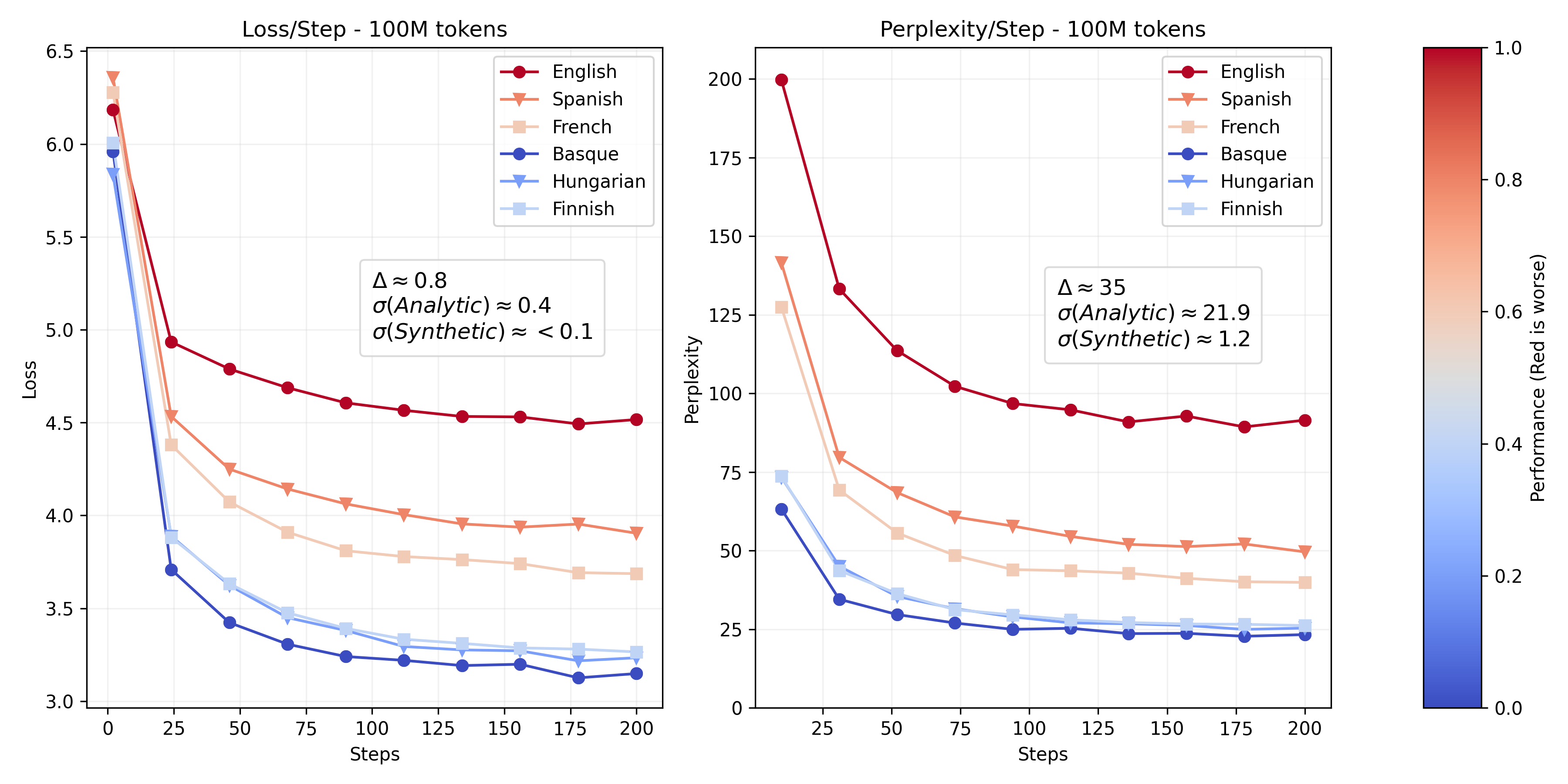}
        \caption{}
        \label{modeling_ind}
    \end{subfigure}
    \begin{subfigure}[b]{0.49\textwidth}
        \includegraphics[width=1\linewidth]{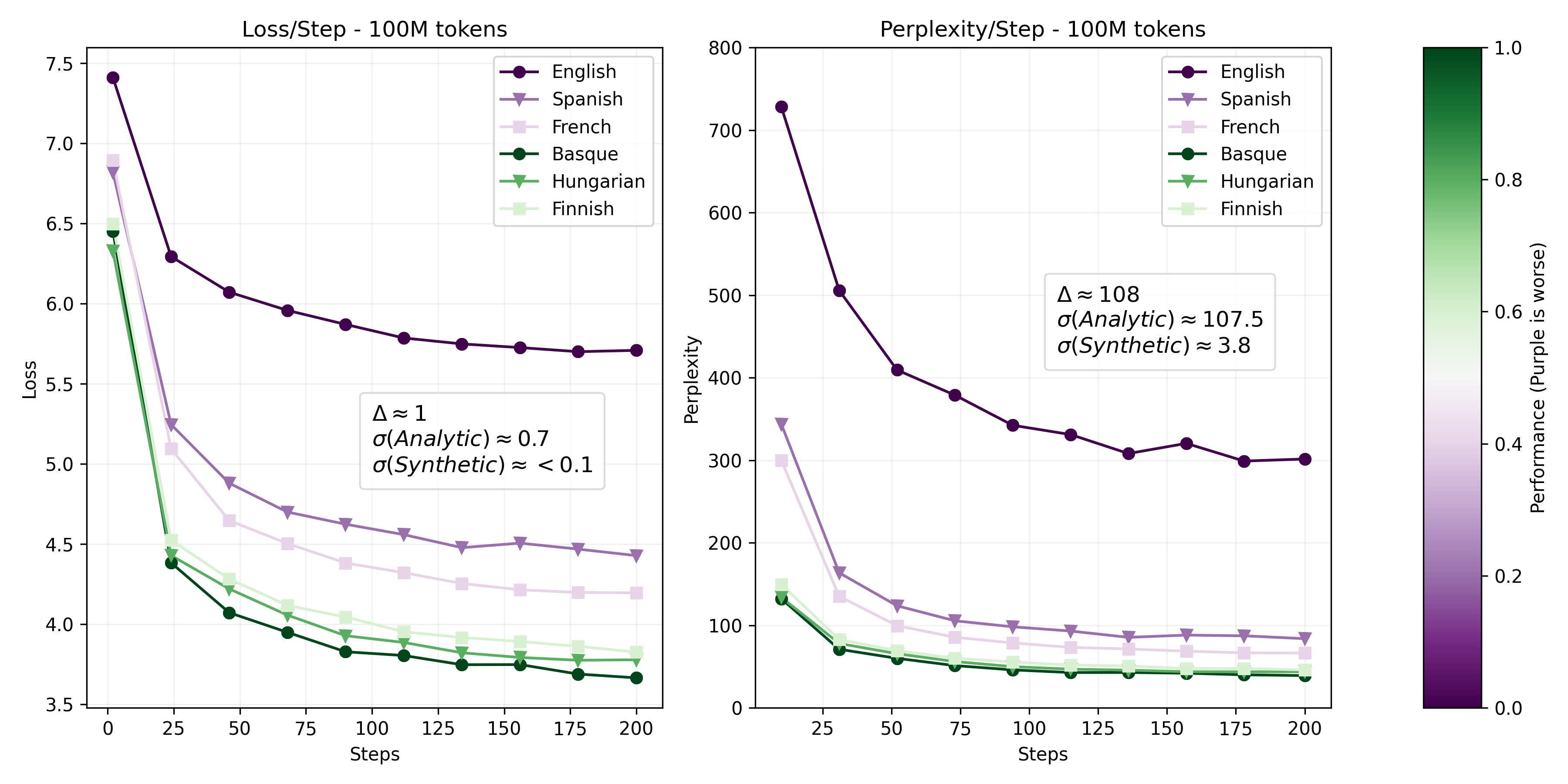}
        \caption{}
        \label{modeling-par}
    \end{subfigure}
    \caption{Results of the training on independent corpora extracted from LCC (a) and validation in PBC corpora (b). Overall, synthetic languages performed better than their analytic counterparts. This is evidenced by lower and more consistent values.}
    \label{both-graphs}
\end{figure*}

Figure \ref{modeling_ind} illustrates the progression of perplexity and loss throughout the training process in independent corpora extracted from the LCC (100M tokens per language). Notably, synthetic languages consistently exhibited superior performance, as indicated by lower loss and perplexity metrics. The average loss for synthetic languages was significantly lower than that of analytic languages. The mean comparison showed a one-point difference in favor of the synthetic group. Additionally, synthetic languages demonstrated greater robustness and compactness ($\sigma<0.1$) compared to analytic languages ($\sigma \approx 0.7$). This suggests that the performance of synthetic languages was not only better but also more stable across different instances. 

In terms of the individual differences, Basque, Hungarian, and Finnish behaved similarly. Basque and Hungarian performed slightly better than the primarily agglutinative Finnish. However, these differences were not significant and practically they performed equally. Interestingly, the results observed in the previous experiments emerged more clearly in the analytic languages. The productivity scale mapped directly in language modeling performance for English, Spanish, and French.

The trends persisted when models were trained on parallel corpora (\ref{modeling-par}). Synthetic languages achieved better generalization more rapidly than analytic languages. The individual differences also held in this experiment. The enhanced generalization ability of Basque, Hungarian, and Finnish may reflect their structural advantage, which enables models to streamline learned patterns more effectively. We hypothesize that the regularity and automaticity of synthetic languages —their predictability— played a relevant role in these results. Regular sequences enhanced the model's ability to predict forms that had not been previously observed. By adhering to consistent and predictable patterns, the model could make more accurate predictions about the characteristics and features of unseen forms. This regularity allowed the model to generalize better from the training to new, unseen data.

\section{Conclusion}
This study examined the impact of morphological typology on tokenization and language modeling, focusing on different synthetic and analytical languages. Our findings indicate that synthetic languages, with their regular and productive morphological typology, significantly benefit from BPE tokenization. The results suggest that this advantage may arise from BPE’s effective handling of complex morphological structures, potentially enhancing model performance and generalization. In contrast, analytic languages presented more significant challenges for BPE, resulting in less efficient tokenization and higher variability. Our results from individual comparisons reflect that the typological continuum from linguistic theory reflects in a variety of experiments.

\section{Limitations and Future Work}
This study's scope is limited to six languages. Additionally, small transformer models were used, which might not fully represent the dynamics observed in larger models. The model evaluation relied on loss and perplexity, which are valuable indicators; however, incorporating additional metrics could provide a more comprehensive understanding of the effects under study. Future research should expand to include a more diverse set of languages, enhancing the generalizability of the findings. Investigating the effects of BPE tokenization with larger models across different languages could provide deeper insights.

\section{Ethical Statement}
All datasets used are publicly available and widely used in the research community. This study did not involve personal or sensitive data, ensuring compliance with privacy regulations. We acknowledge the importance of linguistic diversity and aim to contribute to the understanding and development of equitable language processing tools. Additionally, we recognize the role of factors like vocabulary size or cultural preferences in subword repetition apart from morphological structure. The author declares no conflicts of interest.


\bibliography{custom}
\bibliographystyle{apalike}


\newpage
\section*{NeurIPS Paper Checklist}

\begin{enumerate}

\item {\bf Claims}
    \item[] Question: Do the main claims made in the abstract and introduction accurately reflect the paper's contributions and scope?
    \item[] Answer: \answerYes{} 
    \item[] Justification: The experiments conducted show significant results in favor of the hypothesis.
    \item[] Guidelines:
    \begin{itemize}
        \item The answer NA means that the abstract and introduction do not include the claims made in the paper.
        \item The abstract and/or introduction should clearly state the claims made, including the contributions made in the paper and important assumptions and limitations. A No or NA answer to this question will not be perceived well by the reviewers. 
        \item The claims made should match theoretical and experimental results, and reflect how much the results can be expected to generalize to other settings. 
        \item It is fine to include aspirational goals as motivation as long as it is clear that these goals are not attained by the paper. 
    \end{itemize}

\item {\bf Limitations}
    \item[] Question: Does the paper discuss the limitations of the work performed by the authors?
    \item[] Answer: \answerYes{} 
    \item[] Justification: There is a section for limitations.
    \item[] Guidelines:
    \begin{itemize}
        \item The answer NA means that the paper has no limitation while the answer No means that the paper has limitations, but those are not discussed in the paper. 
        \item The authors are encouraged to create a separate "Limitations" section in their paper.
        \item The paper should point out any strong assumptions and how robust the results are to violations of these assumptions (e.g., independence assumptions, noiseless settings, model well-specification, asymptotic approximations only holding locally). The authors should reflect on how these assumptions might be violated in practice and what the implications would be.
        \item The authors should reflect on the scope of the claims made, e.g., if the approach was only tested on a few datasets or with a few runs. In general, empirical results often depend on implicit assumptions, which should be articulated.
        \item The authors should reflect on the factors that influence the performance of the approach. For example, a facial recognition algorithm may perform poorly when image resolution is low or images are taken in low lighting. Or a speech-to-text system might not be used reliably to provide closed captions for online lectures because it fails to handle technical jargon.
        \item The authors should discuss the computational efficiency of the proposed algorithms and how they scale with dataset size.
        \item If applicable, the authors should discuss possible limitations of their approach to address problems of privacy and fairness.
        \item While the authors might fear that complete honesty about limitations might be used by reviewers as grounds for rejection, a worse outcome might be that reviewers discover limitations that aren't acknowledged in the paper. The authors should use their best judgment and recognize that individual actions in favor of transparency play an important role in developing norms that preserve the integrity of the community. Reviewers will be specifically instructed to not penalize honesty concerning limitations.
    \end{itemize}

\item {\bf Theory Assumptions and Proofs}
    \item[] Question: For each theoretical result, does the paper provide the full set of assumptions and a complete (and correct) proof?
    \item[] Answer: \answerYes{} 
    \item[] Justification: We provide experiments that lead to the theoretical result.
    \item[] Guidelines:
    \begin{itemize}
        \item The answer NA means that the paper does not include theoretical results. 
        \item All the theorems, formulas, and proofs in the paper should be numbered and cross-referenced.
        \item All assumptions should be clearly stated or referenced in the statement of any theorems.
        \item The proofs can either appear in the main paper or the supplemental material, but if they appear in the supplemental material, the authors are encouraged to provide a short proof sketch to provide intuition. 
        \item Inversely, any informal proof provided in the core of the paper should be complemented by formal proofs provided in appendix or supplemental material.
        \item Theorems and Lemmas that the proof relies upon should be properly referenced. 
    \end{itemize}

    \item {\bf Experimental Result Reproducibility}
    \item[] Question: Does the paper fully disclose all the information needed to reproduce the main experimental results of the paper to the extent that it affects the main claims and/or conclusions of the paper (regardless of whether the code and data are provided or not)?
    \item[] Answer: \answerYes{} 
    \item[] Justification: The code will be made available after review. The training includes seeding to maximize reproducibility. 
    \item[] Guidelines:
    \begin{itemize}
        \item The answer NA means that the paper does not include experiments.
        \item If the paper includes experiments, a No answer to this question will not be perceived well by the reviewers: Making the paper reproducible is important, regardless of whether the code and data are provided or not.
        \item If the contribution is a dataset and/or model, the authors should describe the steps taken to make their results reproducible or verifiable. 
        \item Depending on the contribution, reproducibility can be accomplished in various ways. For example, if the contribution is a novel architecture, describing the architecture fully might suffice, or if the contribution is a specific model and empirical evaluation, it may be necessary to either make it possible for others to replicate the model with the same dataset, or provide access to the model. In general. releasing code and data is often one good way to accomplish this, but reproducibility can also be provided via detailed instructions for how to replicate the results, access to a hosted model (e.g., in the case of a large language model), releasing of a model checkpoint, or other means that are appropriate to the research performed.
        \item While NeurIPS does not require releasing code, the conference does require all submissions to provide some reasonable avenue for reproducibility, which may depend on the nature of the contribution. For example
        \begin{enumerate}
            \item If the contribution is primarily a new algorithm, the paper should make it clear how to reproduce that algorithm.
            \item If the contribution is primarily a new model architecture, the paper should describe the architecture clearly and fully.
            \item If the contribution is a new model (e.g., a large language model), then there should either be a way to access this model for reproducing the results or a way to reproduce the model (e.g., with an open-source dataset or instructions for how to construct the dataset).
            \item We recognize that reproducibility may be tricky in some cases, in which case authors are welcome to describe the particular way they provide for reproducibility. In the case of closed-source models, it may be that access to the model is limited in some way (e.g., to registered users), but it should be possible for other researchers to have some path to reproducing or verifying the results.
        \end{enumerate}
    \end{itemize}

\item {\bf Open access to data and code}
    \item[] Question: Does the paper provide open access to the data and code, with sufficient instructions to faithfully reproduce the main experimental results, as described in supplemental material?
    \item[] Answer: \answerYes{} 
    \item[] Justification: As mentioned before.
    \item[] Guidelines:
    \begin{itemize}
        \item The answer NA means that paper does not include experiments requiring code.
        \item Please see the NeurIPS code and data submission guidelines (\url{https://nips.cc/public/guides/CodeSubmissionPolicy}) for more details.
        \item While we encourage the release of code and data, we understand that this might not be possible, so “No” is an acceptable answer. Papers cannot be rejected simply for not including code, unless this is central to the contribution (e.g., for a new open-source benchmark).
        \item The instructions should contain the exact command and environment needed to run to reproduce the results. See the NeurIPS code and data submission guidelines (\url{https://nips.cc/public/guides/CodeSubmissionPolicy}) for more details.
        \item The authors should provide instructions on data access and preparation, including how to access the raw data, preprocessed data, intermediate data, and generated data, etc.
        \item The authors should provide scripts to reproduce all experimental results for the new proposed method and baselines. If only a subset of experiments are reproducible, they should state which ones are omitted from the script and why.
        \item At submission time, to preserve anonymity, the authors should release anonymized versions (if applicable).
        \item Providing as much information as possible in supplemental material (appended to the paper) is recommended, but including URLs to data and code is permitted.
    \end{itemize}

\item {\bf Experimental Setting/Details}
    \item[] Question: Does the paper specify all the training and test details (e.g., data splits, hyperparameters, how they were chosen, type of optimizer, etc.) necessary to understand the results?
    \item[] Answer: \answerYes{} 
    \item[] Justification: We add training information and model details on the language modeling experiment section.
    \item[] Guidelines:
    \begin{itemize}
        \item The answer NA means that the paper does not include experiments.
        \item The experimental setting should be presented in the core of the paper to a level of detail that is necessary to appreciate the results and make sense of them.
        \item The full details can be provided either with the code, in appendix, or as supplemental material.
    \end{itemize}

\item {\bf Experiment Statistical Significance}
    \item[] Question: Does the paper report error bars suitably and correctly defined or other appropriate information about the statistical significance of the experiments?
    \item[] Answer: \answerYes{} 
    \item[] Justification: This is added right before the result is presented.
    \item[] Guidelines:
    \begin{itemize}
        \item The answer NA means that the paper does not include experiments.
        \item The authors should answer "Yes" if the results are accompanied by error bars, confidence intervals, or statistical significance tests, at least for the experiments that support the main claims of the paper.
        \item The factors of variability that the error bars are capturing should be clearly stated (for example, train/test split, initialization, random drawing of some parameter, or overall run with given experimental conditions).
        \item The method for calculating the error bars should be explained (closed form formula, call to a library function, bootstrap, etc.)
        \item The assumptions made should be given (e.g., Normally distributed errors).
        \item It should be clear whether the error bar is the standard deviation or the standard error of the mean.
        \item It is OK to report 1-sigma error bars, but one should state it. The authors should preferably report a 2-sigma error bar than state that they have a 96\% CI, if the hypothesis of Normality of errors is not verified.
        \item For asymmetric distributions, the authors should be careful not to show in tables or figures symmetric error bars that would yield results that are out of range (e.g. negative error rates).
        \item If error bars are reported in tables or plots, The authors should explain in the text how they were calculated and reference the corresponding figures or tables in the text.
    \end{itemize}

\item {\bf Experiments Compute Resources}
    \item[] Question: For each experiment, does the paper provide sufficient information on the computer resources (type of compute workers, memory, time of execution) needed to reproduce the experiments?
    \item[] Answer: \answerYes{} 
    \item[] Justification: Provided in the Language Modeling section
    \item[] Guidelines:
    \begin{itemize}
        \item The answer NA means that the paper does not include experiments.
        \item The paper should indicate the type of compute workers CPU or GPU, internal cluster, or cloud provider, including relevant memory and storage.
        \item The paper should provide the amount of compute required for each of the individual experimental runs as well as estimate the total compute. 
        \item The paper should disclose whether the full research project required more compute than the experiments reported in the paper (e.g., preliminary or failed experiments that didn't make it into the paper). 
    \end{itemize}
    
\item {\bf Code Of Ethics}
    \item[] Question: Does the research conducted in the paper conform, in every respect, with the NeurIPS Code of Ethics \url{https://neurips.cc/public/EthicsGuidelines}?
    \item[] Answer: \answerYes{} 
    \item[] Justification: There is an ethics statement section.
    \item[] Guidelines:
    \begin{itemize}
        \item The answer NA means that the authors have not reviewed the NeurIPS Code of Ethics.
        \item If the authors answer No, they should explain the special circumstances that require a deviation from the Code of Ethics.
        \item The authors should make sure to preserve anonymity (e.g., if there is a special consideration due to laws or regulations in their jurisdiction).
    \end{itemize}

\item {\bf Broader Impacts}
    \item[] Question: Does the paper discuss both potential positive societal impacts and negative societal impacts of the work performed?
    \item[] Answer: \answerYes{} 
    \item[] Justification: Briefly mentioned in the Ethical Statement
    \item[] Guidelines:
    \begin{itemize}
        \item The answer NA means that there is no societal impact of the work performed.
        \item If the authors answer NA or No, they should explain why their work has no societal impact or why the paper does not address societal impact.
        \item Examples of negative societal impacts include potential malicious or unintended uses (e.g., disinformation, generating fake profiles, surveillance), fairness considerations (e.g., deployment of technologies that could make decisions that unfairly impact specific groups), privacy considerations, and security considerations.
        \item The conference expects that many papers will be foundational research and not tied to particular applications, let alone deployments. However, if there is a direct path to any negative applications, the authors should point it out. For example, it is legitimate to point out that an improvement in the quality of generative models could be used to generate deepfakes for disinformation. On the other hand, it is not needed to point out that a generic algorithm for optimizing neural networks could enable people to train models that generate Deepfakes faster.
        \item The authors should consider possible harms that could arise when the technology is being used as intended and functioning correctly, harms that could arise when the technology is being used as intended but gives incorrect results, and harms following from (intentional or unintentional) misuse of the technology.
        \item If there are negative societal impacts, the authors could also discuss possible mitigation strategies (e.g., gated release of models, providing defenses in addition to attacks, mechanisms for monitoring misuse, mechanisms to monitor how a system learns from feedback over time, improving the efficiency and accessibility of ML).
    \end{itemize}
    
\item {\bf Safeguards}
    \item[] Question: Does the paper describe safeguards that have been put in place for responsible release of data or models that have a high risk for misuse (e.g., pretrained language models, image generators, or scraped datasets)?
    \item[] Answer: \answerNA{} 
    \item[] Justification: \answerNA{}
    \item[] Guidelines:
    \begin{itemize}
        \item The answer NA means that the paper poses no such risks.
        \item Released models that have a high risk for misuse or dual-use should be released with necessary safeguards to allow for controlled use of the model, for example by requiring that users adhere to usage guidelines or restrictions to access the model or implementing safety filters. 
        \item Datasets that have been scraped from the Internet could pose safety risks. The authors should describe how they avoided releasing unsafe images.
        \item We recognize that providing effective safeguards is challenging, and many papers do not require this, but we encourage authors to take this into account and make a best faith effort.
    \end{itemize}

\item {\bf Licenses for existing assets}
    \item[] Question: Are the creators or original owners of assets (e.g., code, data, models), used in the paper, properly credited and are the license and terms of use explicitly mentioned and properly respected?
    \item[] Answer: \answerYes{} 
    \item[] Justification: Data and code are open source and are cited in the paper.
    \item[] Guidelines:
    \begin{itemize}
        \item The answer NA means that the paper does not use existing assets.
        \item The authors should cite the original paper that produced the code package or dataset.
        \item The authors should state which version of the asset is used and, if possible, include a URL.
        \item The name of the license (e.g., CC-BY 4.0) should be included for each asset.
        \item For scraped data from a particular source (e.g., website), the copyright and terms of service of that source should be provided.
        \item If assets are released, the license, copyright information, and terms of use in the package should be provided. For popular datasets, \url{paperswithcode.com/datasets} has curated licenses for some datasets. Their licensing guide can help determine the license of a dataset.
        \item For existing datasets that are re-packaged, both the original license and the license of the derived asset (if it has changed) should be provided.
        \item If this information is not available online, the authors are encouraged to reach out to the asset's creators.
    \end{itemize}

\item {\bf New Assets}
    \item[] Question: Are new assets introduced in the paper well documented and is the documentation provided alongside the assets?
    \item[] Answer: \answerNA{} 
    \item[] Justification: \answerNA{}
    \item[] Guidelines:
    \begin{itemize}
        \item The answer NA means that the paper does not release new assets.
        \item Researchers should communicate the details of the dataset/code/model as part of their submissions via structured templates. This includes details about training, license, limitations, etc. 
        \item The paper should discuss whether and how consent was obtained from people whose asset is used.
        \item At submission time, remember to anonymize your assets (if applicable). You can either create an anonymized URL or include an anonymized zip file.
    \end{itemize}

\item {\bf Crowdsourcing and Research with Human Subjects}
    \item[] Question: For crowdsourcing experiments and research with human subjects, does the paper include the full text of instructions given to participants and screenshots, if applicable, as well as details about compensation (if any)? 
    \item[] Answer: \answerNA{} 
    \item[] Justification: \answerNA{}
    \item[] Guidelines:
    \begin{itemize}
        \item The answer NA means that the paper does not involve crowdsourcing nor research with human subjects.
        \item Including this information in the supplemental material is fine, but if the main contribution of the paper involves human subjects, then as much detail as possible should be included in the main paper. 
        \item According to the NeurIPS Code of Ethics, workers involved in data collection, curation, or other labor should be paid at least the minimum wage in the country of the data collector. 
    \end{itemize}

\item {\bf Institutional Review Board (IRB) Approvals or Equivalent for Research with Human Subjects}
    \item[] Question: Does the paper describe potential risks incurred by study participants, whether such risks were disclosed to the subjects, and whether Institutional Review Board (IRB) approvals (or an equivalent approval/review based on the requirements of your country or institution) were obtained?
    \item[] Answer: \answerNA{} 
    \item[] Justification: \answerNA{}
    \item[] Guidelines:
    \begin{itemize}
        \item The answer NA means that the paper does not involve crowdsourcing nor research with human subjects.
        \item Depending on the country in which research is conducted, IRB approval (or equivalent) may be required for any human subjects research. If you obtained IRB approval, you should clearly state this in the paper. 
        \item We recognize that the procedures for this may vary significantly between institutions and locations, and we expect authors to adhere to the NeurIPS Code of Ethics and the guidelines for their institution. 
        \item For initial submissions, do not include any information that would break anonymity (if applicable), such as the institution conducting the review.
    \end{itemize}

\end{enumerate}

\end{document}